# AUTONODE: A Neuro-Graphic Self-Learnable Engine for Cognitive GUI Automation

Arkajit Datta
SuperAGI Research
arkajit@superagi.com

Tushar Verma
SuperAGI Research
tushar.verma@superagi.com

Rajat Chawla
SuperAGI Research
rajat@superagi.com

Mukunda N.S
SuperAGI Research
mukunda@superagi.com

Ishaan Bhola
SuperAGI Research
ishaan@superagi.com

*Abstract*— In recent advancements within the domain of Large Language Models (LLMs), there has been a notable emergence of agents capable of addressing Robotic Process Automation (RPA) challenges through enhanced cognitive capabilities and sophisticated reasoning. This development heralds a new era of scalability and human-like adaptability in goal attainment. In this context, we introduce AUTONODE (Autonomous User-interface Transformation through Online Neuro-graphic Operations and Deep Exploration). AUTONODE employs advanced neuro-graphical techniques to facilitate autonomous navigation and task execution on web interfaces, thereby obviating the necessity for predefined scripts or manual intervention. Our engine empowers agents to comprehend and implement complex workflows, adapting to dynamic web environments with unparalleled efficiency. Our methodology synergizes cognitive functionalities with robotic automation, endowing AUTONODE with the ability to learn from experience. We have integrated an exploratory module, DoRA (*Discovery and mapping Operation for graph Retrieval Agent*), which is instrumental in constructing a knowledge graph that the engine utilizes to optimize its actions and achieve objectives with minimal supervision. The versatility and efficacy of AUTONODE are demonstrated through a series of experiments, highlighting its proficiency in managing a diverse array of web-based tasks, ranging from data extraction to transaction processing.

*Keywords—self-operating computer, generative-ai, transformers, vision-transformers, graphs, reinforcement learning*

## I. INTRODUCTION

The advent of Artificial General Intelligence (AGI) heralds a paradigm shift in computational agency, wherein agents are envisaged to possess the capacity for cognition, comprehension, execution, and goal attainment akin to human intelligence. In this context, robotics has made significant strides, demonstrating the ability to perceive and make decisions autonomously. Extending this capability to agents for visualizing and acting upon tasks could revolutionize the automation of redundant tasks on computer systems. A cognitive approach to Robotic Process Automation (RPA) represents a pivotal advancement in the quest for an optimal agent. In this paper, we introduce AUTONODE, a system designed to address RPA challenges through cognitive methodologies. AUTONODE employs a **multi-expert architecture** to facilitate efficient decision-making for subsequent actions. Through extensive experimentation, we identified limitations in Vision-based Large Language Models (LLMs) concerning accurate grounding. To overcome this, we adopted a hybrid approach leveraging Yolo-V8 and Optical Character Recognition (OCR) technologies, enhancing both efficiency and robustness in grounding. A notable challenge encountered was the presence of spurious content, which could impede the identification of the most appropriate next action. AUTONODE incorporates **D**iscovery and mapping **O**peration for graph **R**etrieval **A**gent (DoRA), a mechanism that trains the system to concentrate on the principal elements of the screen, enabling the LLM to make more informed decisions. DoRA also integrates human feedback in a neuro-symbolic manner, enhancing the system's focus on relevant screen areas and mitigating the issue of spurious content. AUTONODE's architecture is heavily influenced by human imitation, aiming to replicate the way a human would interact with a website. The engine also supports a RAG based memory retrieval system which aims at delivering faster Turn Around Time (TAT) for the tasks already done. The system has been validated for scalability with minimal infrastructure requirements. Empirical results demonstrate that AUTONODE's accuracy surpasses that of many existing self-operating computer architectures, with a precision rate exceeding 85%, thereby providing users with a reliable autonomous decision-making tool. The paper is structured as follows: Section 2 delves into related work, followed by a discussion on methodology and system architecture in subsequent sections. Section 4 elaborates on DoRA and its architectural framework. The paper concludes with a presentation of experimental results, comparative analyses, and final remarks.

## II. RELATED WORK

Recent advancements in the integration of large language models (LLMs) and multimodal models with robotic and computer agents have shown promising results across various applications. Research has demonstrated the potential of using pretrained skills to ground LLMs in real-world robotic tasks, enabling robots to complete complex instructions [1]. In the realm of smartphone applications, a multimodal agent framework called App-Agent has been introduced, which operates apps through simplified actions and learns from human demonstrations [2]. The use of multimodal large language models (MLLMs) has also been explored in instruction-based image editing, resulting in notable improvements in automatic metrics and human evaluation [3]. Additionally, the "OS-COPILOT" framework has been proposed for building generalist computer agents, demonstrating strong generalization and self-improvement capabilities. These studies highlight the potential of LLMs and multimodal models in enhancing agent capabilities across various domains, suggesting a promising direction for future research [4].

## III. METHODOLOGY

The principal objective of our research was directed towards addressing the challenges inherent in Robotic Process Automation (RPA) through a cognitive approach. The ambition was to develop an engine characterized both by robustness and the capacity to undertake intelligent subsequent actions to fulfill its designated tasks. Initially, our endeavors embarked from a simplistic paradigm, progressively advancing our architecture through iterative refinement, primarily driven by the array of issues we encountered during the evaluation phase. This process of evolution in our methodology was critical, as it allowed for the identification and rectification of any deficiencies, thereby enhancing the cognitive capabilities of the system.

In our journey towards achieving this, various open-source projects served as invaluable resources, shedding light on the multitude of challenges that our basic approach was susceptible to. These insights were instrumental in guiding our architectural enhancements, ensuring that our approach remained congruent with the overarching goal of cognitive RPA. Fig. 1 in our paper delineates the foundational architecture of AUTONODE, which has been designated as Version-1. This initial architectural framework laid the groundwork for what would later evolve into a more advanced, cognitive RPA solution. The design philosophy behind Version-1 was grounded in the establishment of a foundational set of capabilities that could be iteratively built upon. By doing so, we aimed to incrementally inject a higher degree of cognitive functionality, enabling AUTONODE to not only interpret and navigate web interfaces autonomously but also to execute a wide array of tasks with increasing sophistication. Central to this architectural evolution was the cognizance of the limitations inherent in our initial model. These limitations served as focal points for continued research and development, guiding the systematic enhancement of AUTONODE's capabilities. The iterative refinement process, underscored by a commitment to both scalable and intelligent automation, ultimately contributed to the emergence of an advanced RPA solution capable of autonomously navigating and operating within dynamic web environments. This evolution from a basic to a more intricate architecture reflects our overarching research trajectory, aiming to bridge the gap between conventional RPA and cognitive automation through the amalgamation of neuro-graphic operations and deep exploration techniques. Through this approach, AUTONODE was transformed into a more adaptive, efficient, and intelligent engine capable of executing complex workflows and adapting to new challenges with minimal human intervention.

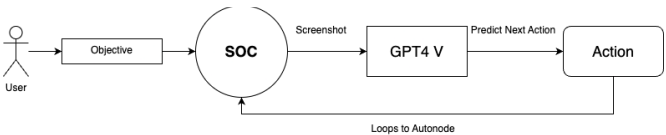

Fig. 1. Basic architecture of AUTONODE – Version: 1

Within the architectural framework of AUTONODE, the system leverages the capabilities of the GPT4-V API to deduce the subsequent optimal action based on visual inputs. This mechanism involves the engine capturing a screenshot of the current state of the computer interface, which serves as the basis for decision-making processes aligned with predefined objectives. Initially, AUTONODE incorporated the PyAutoGUI library to execute actions within the graphical user interface (GUI) environment based on the predictive analytics of the model. Let $S_t$ represent the state of the GUI at time $t$, derived from a screenshot and precoesed into a structured input vector. $S_t$ includes all visual elements currently displayed to the user. The GPT4-V takes $S_t$ as an input with the prompt being $P_t$ produces a decision vector $D_t$, representing the optimal action(s) to be executed at time $t$. This response is passed through parser denoted as $f_{Parser}$. The flow of the architecture is discussed in (1) and (2).

$$D_t = f_{VM}(S_t, P_t) \quad (1)$$

$$D_{proc} = f_{Parser}(D_t) \quad (2)$$

Here, $f_{VM}$ denotes the mapping function realized as vision model to be GPT-4V that transforms current state $S_t$ into the decision vector $D_t$ and eventually to a more parsable output of $D_{proc}$. These actions are enabled through PyAutoGUI library, facilitating the interaction with the GUI. The execution of an action, $A_t$, can be observed in the equation (3).

$$A_t(D_{proc}) = \begin{cases} Click(x,y) & if\ D_{proc}\ \epsilon\ click \\ Type(text) & if\ D_{proc}\ \epsilon\ type \\ Scroll(amount) & if\ D_{proc}\ \epsilon\ scroll \\ Hover(x,y) & if\ D_{proc}\ \epsilon\ hower \end{cases} \quad (3)$$

where $x, y$ represents coordinates for the mouse options and $text$ representing text to type. Here, $A_t$ represents the action taken by AUTONODE from the action space conditioned on the value of $D_{proc}$ to be taken at time $t$. The actions specified in $D_{proc}$ can be a part of the action $A_t$ directly influences the GUI, leading to a transition into a new state $S_{t+1}$, which is subsequently evaluated for further actions, creating a feedback loop for continuous task execution as represented in (4).

$$S_{t+1} = g(S_t, A_t) \quad (4)$$

with $g$ symbolizing the transition function that encapsulates the effect of action $A_t$ on the current state $S_t$ to yield the next state $S_{t+1}$.

The basic architecture had several shortcomings which needed to be fixed in order to achieve to the best architecture for AUTONODE. The shortcomings are given as follows –

- Vision Model gave irrelevant/ wrong location to click.
- There were series of hallucinations recorded in the model. Like going on the wrong element for a particular action. For example: clicking on the search bar rather on the compose email for composing an email.

- Problem of spurious content in the prompt was observed. In this architecture we are passing the whole clickable element's list to the LLM which is making it hallucinate.

- Model was unable to generate the next best action, hence was unable to complete the given task.

*E. Algorithm Selection*

The process of algorithmic optimization necessitates continuous iterations to rectify discovered deficiencies. Within the foundational architecture, a prominent issue identified was the erroneous selection of click locations, significantly exacerbating the system's propensity for inaccuracies. This realization led to the conclusion that the existing Visual Model lacked the requisite capability to accurately identify the appropriate areas for interaction. Consequently, it became imperative to integrate specialized models dedicated to this aspect of the task, ensuring a more precise and reliable system performance.

*Process A*

Fixing the problem with the unpredictable clicking was very important because it was causing delays for other parts of our system. The attainment of comprehensive cognitive automation necessitated prioritizing the rectification of this issue. The foundational approach to resolving the challenge of discerning accurate coordinates for textual elements on web interfaces hinges on the application of Optical Character Recognition (OCR) technology. In conjunction with this technology, we integrated a fine-tuned YOLO (You Only Look Once) model, specifically tailored for the detection of web-based elements. For this purpose, the YOLO-v8-m version was selected for fine-tuning tasks, reflecting our commitment to leveraging advanced models for improved performance. Figure 2 elucidates the initial evolutionary phase of AUTONODE's architecture, illustrating the integration of OCR capabilities. This enhancement marks a pivotal advancement in AUTONODE's ability to navigate and interact with web interfaces autonomously, thereby significantly mitigating previous limitations encountered in identifying and interacting with textual elements.

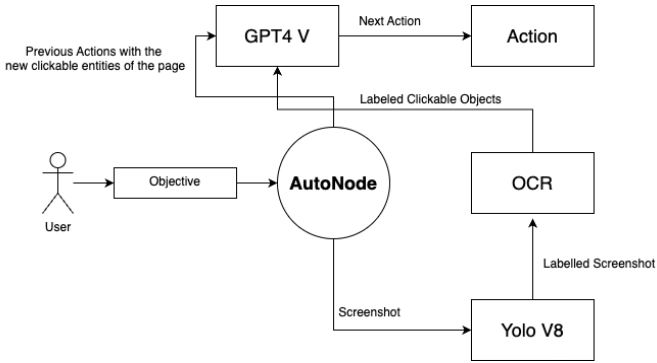

Fig. 2. Process A iteration of AUTONODE with OCR capabilty

Let $S_t$ represent the state of the GUI at time $t$, derived from a screenshot and precoesed into a structured input vector. Unlike the last process the image this time passes through several pre-processing to extract the relevant information. $S_t$ is passed to function $f_{YOLO}$ to receive multiple cropped web element's images. Hence we can say those images are subset of $S_t$ as referred in (5).

$$C_t, L_{C_t} = f_{YOLO}(S_t) \quad (5)$$

Where, $C_t$ is the set of cropped images detected from $S_t$ using the function of YOLO-v8. $L_{C_t}$ is the location of the bounding box also called bbox, this location will be used in the future when AUTONODE decides to click on that element. Now as there are several images as represented in (6). OCR is ran on each of these images to retrieve the text written inside the element, this helps the Vision Model to take a better decision on which element to choose.

$$C_t = \{c_{t,1}, c_{t,2}, c_{t,3}, c_{t,5}, c_{t,6} \ldots \ldots c_{t,n}\} \quad (6)$$

$$T_t = f_{OCR}(c_{t,i}) \; \forall \, i \in \{1, n\} \quad (7)$$

$$T_t = \{t_{t,1}, t_{t,2}, t_{t,3}, t_{t,5}, t_{t,6} \ldots \ldots t_{t,n}\} \quad (8)$$

Here, $T_t$ is the textual data extracted from the images $C_t$ using the OCR function $f_{OCR}$. Now, all textual data with their respective locations are passed to vision model function $f_{VM}$ to produce the decision vector $D_t$ as represented in (9).

$$D_t = f_{VM}(S_t, t_{t,i}, L_{c_{t,i}}) \; \forall \, i \in \{1, n\} \quad (9)$$

Now, as represented in (2), (3) and (4) decision vector $D_t$ is used to take the action $A_t$ which in turn changes the state of GUI from $S_t$ to $S_{t+1}$. An enhancement introduced involves the documentation of actions undertaken thus far, which are subsequently relayed to the vision model. Given the formidable capability of Large Language Models (LLMs) to discern the extent of task completion, they adeptly select the most optimal subsequent action. This process not only leverages the inherent understanding and processing power of LLMs but also contributes to a more efficient and informed decision-making mechanism within the system. By integrating this continuity of action recognition and selection, the model significantly improves in task execution efficiency, enabling a more seamless progression towards goal attainment. Hence a variable $H_t$ (History) is initailised which concatetates all the actions taken from the beginning of the automation as shown in (10).

$$H_t = A_{t,i} \; \forall \, i \in \{1, n\} \quad (10)$$

Consequently, we can reformulate equation (9) to represent the final structural composition of Process A, as depicted in equation (11).

$$D_t = f_{VM}(S_t, t_{t,i}, L_{c_{t,i}}, H_t) \; \forall \, i \in \{1, n\} \quad (11)$$

Comprehensive evaluations across diverse scenarios have revealed that the enhancements introduced in AUTONODE, while substantially beneficial, are not without their challenges. A notable anomaly identified is the incidence of action hallucination, a phenomenon where the Large Language Model (LLM) within the Vision Model component inaccurately predicts the subsequent operational step. This discrepancy arises from the model's limited contextual awareness regarding the intricacies of navigating web interfaces. Given the substantial volume of data presented on a typical web screen, the LLM is predisposed to 'hallucinate' or erroneously identify an incorrect course of action. For instance, when tasked with filtering unread emails within a Gmail interface, the model lacks the explicit knowledge required to interact with the search bar and input the specific command "unread: label" to filter unread messages. Consequently, in complex scenarios as described, the model's performance is observed to deviate from expected accuracies, underscoring a critical area for further refinement. This leads to the next iteration of AUTONODE.

Process B

Addressing the challenge of determining the subsequent optimal action can be systematically approached by provisioning Large Language Models (LLMs) with explicit knowledge concerning the requisite steps for task completion. To this end, our experimentation involved imparting contextual information as guidance to the LLM, a strategy that substantively enhanced the engine's operational effectiveness. The architectural framework underpinning this approach, designated as Process B, is elucidated in Fig. 3. This methodology underlies our efforts to refine the decision-making capabilities of LLMs, thereby optimizing their performance in executing designated tasks with greater precision and efficiency.

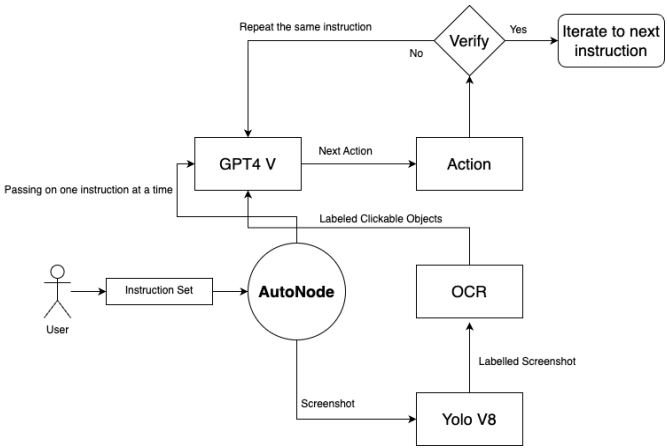

Fig. 3. Process B iteration of AUTONODE with Instruction set and Verification module.

In this process the basic processing steps are same in YOLO and OCR what changes is giving the instruction steps to the LLM for it to take a valid next action. Hence we can rewrite (11) with the new iteration as shown in (12). Let's assume $I_t$ is the Instruction set at time $t$.

$$D_t = f_{VM}\left(S_t, t_{t,i}, L_{c_{t,i}}, H_t, I_t\right) \forall\, i \in \{1, n\} \quad (12)$$

In this way we have observed a significant reduction in hallucination and wrong next action prediction. Another update in this version was a verification module $f_{verify}$ which can assist the engine from taking any wrong decisions, as wrong actions in this environment can be very costly. The verification module takes everything with the updated screenshot with a mark on the area of clicking represented as $S_{t,ROI}$ and action to be taken shown as $A_t$.

$$A_{verified} = f_{verfiy}\left(S_{t,ROI}, t_{t,i}, L_{c_{t,i}}, H_t, I_t, A_t\right) \forall\, i \in \{1, n\} \quad (13)$$

If $A_{verified}$ returns a positive result the engine moves ahead otherwise it loops on completing the current action correctly by re-iterating on getting an update action. These updates had improvements yet added issues in the User Experience (UX) as the person using the engine won't be comfortable enough to give the whole set of instructions. Also, the issue of spurious content was still intact where the model hallucinates while choosing which element to click. Solving these problems lead us to develop our final iteration Process C.

Process C

In the course of interacting with digital interfaces, human users inherently prioritize attention to specific regions of the screen, effectively filtering out extraneous information. This behavioral pattern suggests a strategic approach to refining the input provided to Large Language Models (LLMs) within the context of cognitive Robotic Process Automation (RPA). By analyzing data derived from numerous workflow instances, it is feasible to discern and delineate key areas or Regions of Interest (ROI) on the interface that warrant focused engagement. An advanced methodology can be established by cataloging these ROIs in relation to the preceding interactive element. Consequently, this facilitates the construction of a hierarchical data structure, wherein a parent node signifies the element requiring activation, and its child nodes represent subsequent ROIs of relevance.

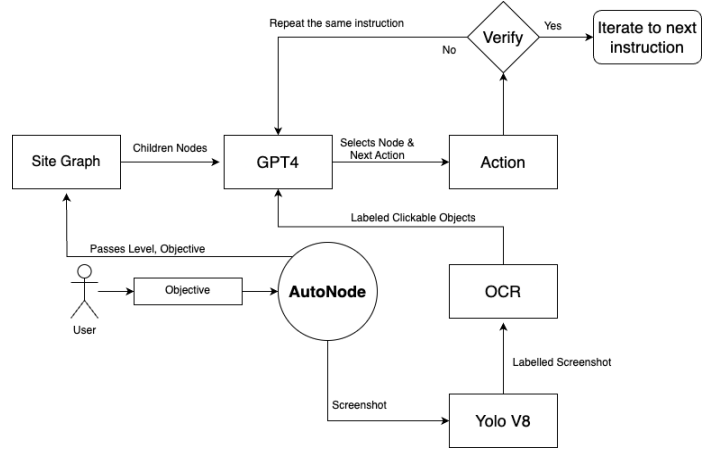

Fig. 4. Process C iteration with neuro-graph based architecture

Such a paradigm shift significantly mitigates the incidence of inaccuracies and the generation of irrelevant actions by the LLM, attributable to the deliberate exclusion of non-essential content from the processing spectrum. Furthermore, navigating this structured graph ostensibly obviates the need for a Vision Model, thereby markedly decreasing the overall operational expenses associated with the system. This innovative approach, whose architecture is shown in Fig. 4 underscores a pivotal enhancement in the efficiency and reliability of cognitive automation tasks, paving the way for more effective utilization of LLM capabilities in RPA solutions.

---

**Algorithm 1** AutoNode: Process C ( w/ Neuro-Graphic Site Architecture )

---
**Require:** Site Graph $G$, User Objective $O$, Vision Model $V$ (YOLOv8 + OCR), LLM (GPT-4)
**Ensure:** Successful interaction with the digital interface based on similarity metrics
1: **function** CALCULATEJAROSIMILARITY($str1$, $str2$)
2:    $m \leftarrow$ Number of matching characters in $str1$ and $str2$
3:    $t \leftarrow$ Number of transpositions divided by 2
4:    $sim_j \leftarrow \frac{1}{3}\left(\frac{m}{|str1|} + \frac{m}{|str2|} + \frac{m-t}{m}\right)$
5:    **return** $sim_j$
6: **end function**
7: **function** FINDCHILDRENNODES($PrevNode$)
8:    **return** CHILDRENNODESFUNCTION($PrevNode$)
9: **end function**
10: **function** SELECTNODE($ChildrenNodes$, $Objective$, $History$)
11:    **return** LLMDECISIONFUNCTION($ChildrenNodes, Objective, History$)
12: **end function**
13: **function** GETELEMENTLOCATION($NodeSelected$)
14:    $T, L \leftarrow$ Use YOLOv8 and OCR to find $NodeSelected$ location
15:    **return** $T, L$
16: **end function**
17: **function** GETSIMILARITYSCORE($NodeSelected$, $T$, $L$)
18:    $f_{semantic} \leftarrow$ CALCULATEJAROSIMILARITY(NodeSelected.Name, $T$)
19:    $f_{euclidean} \leftarrow$ CalculateEuclideanDistance($NodeSelected.Location$, $L$)
20:    $similarity \leftarrow 0.7 \times f_{semantic} + 0.3 \times f_{euclidean}$
21:    **return** $similarity$
22: **end function**
23: **function** AUTONODEV3($G$, $O$)
24:    $History \leftarrow$ Initialize or retrieve previous history
25:    $CurrentNode \leftarrow$ Initial node of $G$
26:    **while** Objective not met **do**
27:      $ChildrenNodes \leftarrow$ FINDCHILDRENNODES($CurrentNode$)
28:      $NodeSelected \leftarrow$ SELECTNODE($ChildrenNodes$, $O$, $History$)
29:      $T, L \leftarrow$ GETELEMENTLOCATION($NodeSelected$)
30:      $similarity \leftarrow$ GETSIMILARITYSCORE($NodeSelected$, $T$, $L$)
31:      **if** $similarity$ is highest among $ChildrenNodes$ **then**
32:         Take action on $NodeSelected$
33:      **else**
34:         Go back to previous node or repeat selection
35:      **end if**
36:      $History \leftarrow History + \{NodeSelected\}$
37:      $CurrentNode \leftarrow NodeSelected$
38:    **end while**
39: **end function**
40: AUTONODEV3($SiteGraph, UserObjective$)

---

The concept of Neuro-Graphic site architecture embodies a graph-based framework for cataloging information pertinent to any website's structure. To elucidate this concept, consider the operation of sending an email via Gmail, starting from the Google homepage. Initially, the user or an automated agent must select the 'Gmail' option located on the top right corner, thereby navigating to the Gmail interface. Within this context, it is critical for a bot (designed to send and reply to emails) to prioritize regions of interest (ROIs) on the webpage, which, in this scenario, include the 'Compose' button, the 'Search in mail' bar, and the visible emails. Subsequently, upon selecting 'Compose,' the focus narrows to a newly opened dialog within which the ensuing ROIs are sequentially the 'To' text field, followed by 'Subject' and 'Body' fields, culminating in the 'Send' button. This linear progression through singular child nodes mirrors human navigation patterns, enhancing the system's decision-making robustness. Illustrated in Figure 5 is a neuro-graphic representation tailored for the use case of automating email composition and responses. Displayed within are nodes representing emails labeled as Mail1, Mail2, and Mail3, indicative of emails visible on the interface under a hypothetical scenario. The presence of multiple singular child nodes throughout this representation serves to streamline the focus onto individual nodes, thereby eschewing the need to scan the entire screen. This neuro-graphic tree employs a greedy Depth First Search (DFS) approach for traversal, with the Large Language Model (LLM) serving as the arbiter at each juncture. This integration of neuro-graphic site architecture and LLM decision-making mechanisms underscores an innovative approach towards enhancing the efficiency and robustness of performing web-based tasks, such as email management, through automated systems.

Fig. 5. Gmail Neuro-graph for a compose and reply automation engine

The process of YOLO and OCR is similar as it is being done in the earlier versions. The additional step is to parse throught the site tree to find the children nodes and then use the LLM to traverse to the next node. Let's say $f_{traverse}$ is the function which gives the children nodes. These children nodes are then passed to $f_{LLM}$ for it to decide on the which node to traverse to. Action is then taken on the node selected. There is also a verification loop which helps in deciding whether the node selected is the correct element to select or not. In (14) $PrevNode_t$ is the previous nodes of the graph at time t. In (15) $Nodes_t$ represents all the children nodes which have to be taken in consideration for taking the actions. These nodes with the History and the Objectives is passed to the LLM to find the node selected as shown in (16).

$$Nodes_t = f_{traverse}(PrevNode_t) \quad (14)$$

$$Nodes_t = \{N_{t,1}, N_{t,2}, N_{t,3}, N_{t,4} \ldots \ldots, N_{t,n}\} \quad (15)$$

$$Node_{selected} = f_{LLM}(N_{t,i}, Objective, H_t) \; \forall \; i \in \{1, n\} \quad (16)$$

The Node Selected contains some meta-data about various parameters related to the element which has to be clicked. The location of the element is still calculated using YOLO and OCR hence using (8) we retrieve $T_t$ and $L_{C_t}$ of the elements present in the screenshot $S_t$. The metadata of Node selected contains the reference location with the name of element to be clicked, a semantic similarity index is calculated using jaro similarity as presented in (17). The equation shown in (18) gives a similarity score on which the highest scored element is picked.

$$sim_j = \frac{1}{3}\left(\frac{m}{|s_1|} + \frac{m}{|s_2|} + \frac{m-t}{m}\right) \quad (17)$$

Here, m is A character from one string is considered matching with a character from the second string only if the characters are the same and their positions do not differ by more than $\left[\frac{\max(|S_1|,|S_2|)}{2}\right] - 1$, where |s1| and |s2| are the lengths of the string. And t is transposition which is, After the matching characters are identified, transpositions are counted. A transposition is considered for each pair of matching characters that are in a different order in the two strings. The total number of transpositions is divided by 2 because this way each transposed character is counted only once.

$$similarity = 0.7 * f_{semantic\,sim}(Node_{selected}, t_{t,i}) + 0.3 * \left(1 - f_{euclidean}(Node_{selected}, L_{C_{t,i}})\right) \; \forall \; i \in \{1, n\} \quad (18)$$

At the end of (18) AUTONODE receives the accurate location to take the action $A_t$. Similar steps of taking actions are performed as discussed in (3).

*E. Validation & Selection*

To ascertain the efficacy and optimize the solution architecture for AUTONODE, a rigorous methodology encompassing testing and iterative enhancement has been employed. This validation process is meticulously designed, leveraging an array of use cases that encapsulate a breadth of scenarios and challenges pertinent to the AUTONODE application domain. Through this comprehensive testing regime, the design and operational paradigms of AUTONODE are refined, ensuring the attainment of the most suitable and effective solution for its intended functionalities.

*E. Iterative Refinement*

During our research, AUTONODE has undergone a series of iterative refinements, culminating in a significantly advanced and mature framework capable of addressing and resolving the various limitations identified earlier. The foundation of AUTONODE is constructed upon a neuro-graphical system, which has demonstrated remarkable proficiency in enhancing both the user experience (UX) and the overall effectiveness of the platform. A pivotal aspect of our methodology is its emphasis on resilience and fault tolerance, thereby ensuring the robustness of the cognitive Robotic Process Automation (RPA) capabilities. Integral to the enhanced version of AUTONODE is a graph-based module, which derives its power from a self-training component, herein referred to as DoRA (Deep Robotic Automation). This module, which will receive a more detailed examination in subsequent sections of the study, is instrumental in augmenting the autonomous functionalities of AUTONODE, further solidifying its position as a pioneering solution in the realm of cognitive RPA.

IV. DORA

In the realm of cognitive process automation, the development of self-training modules that can autonomously explore, learn, and adapt mapping to complex and unchartered environments in the interface via exploiting the vision capabilities of multimodal models is of paramount importance. The DoRA (**D**iscovery and mapping **O**peration for graph **R**etrieval **A**gent) framework is a new way to use different types of data, networked data structures, and learning through trial and error to create a flexible exploration agent.

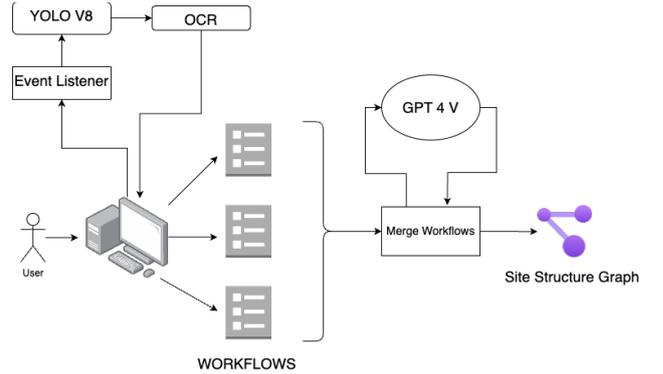

Fig. 6. DoRA Architecture

This section presents an overview of DoRA's methodology, which comprises five integral components aimed at enhancing the cognitive capabilities of automation systems. Firstly, the Guided Exploration module is introduced. This module is designed to navigate through the interface from web applications & desktop applications, identifying relevant information and patterns that can be further processed, laying the foundation for future work on a Generalised Explorer Agent using Reinforcement Learning (RL).

Secondly, the Learnable Mapping and Annotation module is discussed. This component is responsible for establishing meaningful connections between different data elements, facilitating their interpretation and utilisation in various automation flows. By enabling the system to learn and adapt these mappings, DoRA enhances its ability to handle diverse and evolving data structures within interface elements across workflows.

Thirdly, the framework incorporates a Graph-aided Heuristic Search mechanism, which utilises normalised scores to retrieve, encode, and reflect the learnable mappings. This approach ensures that the system can efficiently navigate the site-graph structure, prioritising the most relevant information, minimising computational overhead and translating information into action domain.

Fourthly, DoRA integrates Knowledge Graph Augmented Language Modelling, a technique that leverages knowledge graphs for knowledge-grounded dialogue generation. This component is crucial for grounding and context retrieval from subgraphs, enabling the system to generate more contextually relevant and coherent responses in conversational interfaces.

Finally, the transition from Language Modelling to a Neuro-Symbolic Programming Paradigm is explored. This shift represents a significant advancement in cognitive automation, as it allows for the generalisation of cognitive tasks through multimodal representation learning during training. By combining symbolic reasoning with graph neural network-based learning, DoRA aims to achieve a more holistic and flexible approach to cognitive process automation.

In conclusion, the DoRA framework presents a comprehensive and innovative methodology for self-training module in cognitive process automation. By integrating guided exploration, learnable mapping, graph-aided heuristic search, knowledge graph augmented language modelling, and neurosymbolic programming paradigm, DORA sets a new standard for the development of intelligent automation systems which can be trained on interface intensive workflows. Future research co-laterally will focus on refining these components and exploring their applications in various domains, with the ultimate goal of achieving more autonomous, efficient, and adaptable cognitive processes.

### A. Guided Exploration Module

The Guided Exploration module in the DoRA framework serves as the foundation for autonomous data navigation and pattern identification to model human actions on computer applications across the operating system [5]. This component is crucial for enabling the system to efficiently traverse complex interfaces and extract relevant information and event $e_i$ across the OS environment, hereafter referred to as the 'World Interface', and 'interface 'interchangeably, $W$. At the core of the Exploration module is the concept of guided discovery, where the system is directed to record events of the interface landscape throughout the workflows spanning transformed action steps $a_i$, which need to be interacted with in order to leverage the 'learning' facility of the system, such that

$$W \in (a \subseteq e) \quad (19)$$

where,

$$e = \{e_i \mid i \in W\} \quad (20)$$

$$a = \{a_i \mid a_i = T_{explore}(e_i), i \in W\} \quad (21)$$

The future development of this module involves the integration of Reinforcement Learning (RL) to create a Generalised Explorer Agent in addition to the existing guided discovery abstraction. RL provides a framework for the agent to learn optimal exploration strategies through trial and error, continuously improving its ability to navigate the data environment and make informed decisions about where to focus its attention. Exploration can help an agent gather more information about its environment, which can improve its ability to generalise to new tasks or environments within the interface. This exploration can help agents learn about parts of the environment which may be useful at test time, even if they are not needed for the optimal policy on the training environments. As the context space is sparse, the exploration is guided by a reward function, which quantifies the value of the information discovered during exploration, encouraging the agent to prioritise areas, hereafter called contextual Sample Space, $S_c$ of the interface that are most relevant to the task, $t$ at hand where a task may be composed of many subtasks, $t_i$ which may be employed by the RL agent to derive and abstract the optimal actions, such that for the $j^{th}$ workflow, the subtasks $t_i^j$ constitute the task $t^j$, where action steps $a_i^j$ add up to form the optimal actions, $a^j$ and the action vector, $A^j$.

$$t^j = \{t_i^j \mid i \in W\} \quad (22)$$

The Guided Exploration module is essential for the overall efficiency and effectiveness of the DoRA framework, as it lays the groundwork for subsequent stages of post-processing and analysis. By enabling the system to identify and focus on the most pertinent information, this module enhances the system's ability to adapt to diverse environments and extract actionable insights.

---

**Algorithm 2** Guided Exploration within the DoRA Framework
**Require:** The 'World Interface' $W$, an abstract representation of the OS environment
**Ensure:** Transformed action steps $A$, aligned with corresponding events $E$ in the environment
1: Initialize Event Set $E \leftarrow \{\}$
2: Initialize Action Set $A \leftarrow \{\}$
3: **for** each interface element $i$ in $W$ **do**
4:    Capture event $e_i$ with its metadata:
5:       timestamp $\leftarrow$ GetCurrentTimestamp()
6:       action_type $\leftarrow$ GetActionType($e_i$)
7:       action_description $\leftarrow$ GetActionDescription($e_i$)
8:    Add event metadata to $E$:
9:       $E \leftarrow E \cup \{e_i(\text{timestamp}, \text{action\_type}, \text{action\_description})\}$
10:   Apply transformation function $\tau_{explore}$ to record actions:
11:      $a_i \leftarrow \tau_{explore}(e_i)$
12:   Add transformed action to $A$:
13:      $A \leftarrow A \cup \{a_i\}$
14:   Check if $a_i$ leads to a new state in $W$
15:   **if** a new state is reached **then**
16:      Update $W$ with the new state information
17:   **end if**
18: **end for**
19: **Teach Mode:**
20: **for** each recorded action $a_i$ in $A$ **do**
21:   Present $a_i$ to the user for confirmation
22:   **if** user modifies $a_i$ **then**
23:      Record user's modified action $a_i'$
24:      Update $A$ with $a_i'$:
25:        $A \leftarrow (A \setminus \{a_i\}) \cup \{a_i'\}$
26:   **end if**
27: **end for**
28: **return** $A, E$

---

### B. Learnable Mapping and Annotation Module

The Learnable Mapping and Annotation module is a pivotal component of the DORA framework, responsible for establishing and refining the relationships between different GUI elements. This module enables the system to interpret and organise the information from events and action steps, discovered during the guided exploration phase, facilitating its use in various automation tasks.

Mapping, $M$ in this context refers to the process of linking related data points, creating a structured representation, $S_e$ of the information that can be easily navigated and analysed. In a knowledge graph, mapping involves connecting entities $n$, and their attributes based on their relationships $r$. The learnable aspect of this module implies that these mappings are not static and they can be updated and improved over time as the system encounters new data or as the relationships between data elements evolve.

$$M_i = f_{map}(n_i, r_i) \quad (23)$$
$$S_e = \{M_i \mid i \in W\} \quad (24)$$

---

**Algorithm 3** Learnable Mapping and Annotation in the DoRA Framework
**Require:** The 'World Interface' $W$, a set of events and action steps $S_e$
**Ensure:** Site graph $G$ with structured nodes $S_n$ and mappings $M_i$
1: Initialize Site Graph $G \leftarrow \emptyset$
2: Initialize Structured Information $S_e \leftarrow \{\}$
3: Initialize Mappings $M \leftarrow \{\}$
4: **function** MAPENTITIES($W, S_e$)
5:    **for** each element $i$ in $W$ **do**
6:       Determine entity $n_i$ and relationship $r_i$ from $S_e$
7:       $M_i \leftarrow f_{map}(n_i, r_i)$
8:       $S_e \leftarrow S_e \cup \{M_i\}$
9:    **end for**
10: **end function**
11: **function** ANNOTATEMETADATA($S_e$)
12:    **for** each mapping $M_i$ in $S_e$ **do**
13:       Apply YOLO, OCR, and GPT-4V to annotate $M_i$
14:       $M'_i \leftarrow f_{YOLO, OCR, VM}(M_i)$
15:       Update $S_n$ with $M'_i$:
16:       $S_n \leftarrow_{map}(S_e)$
17:    **end for**
18: **end function**
19: **function** GENERATESITEGRAPH($S_n$)
20:    **for** each structured node $S_n$ **do**
21:       Incorporate $S_n$ into Site Graph $G$
22:    **end for**
23:    **return** $G$
24: **end function**
25: $S_e \leftarrow$ GUIDEDEXPLORATION($W$)    ▷ From previous algorithm
26: $S_n \leftarrow$ ANNOTATEMETADATA($S_e$)
27: $G \leftarrow$ GENERATESITEGRAPH($S_n$)
28: **return** $G$

---

Annotation, on the other hand, involves adding metadata and labels to the data points, providing additional context and categorisation among the GUI elements across the OS. During the post processing regimes, YOLO and Optical Character Recognition and GPT-4V methods are used to annotate the metadata to map the information gathered in the exploration phase, $S_e$ into transformed Structured "Nodes" vectors, $S_n$, within the site-graph, $G$ [11]. The Learnable Mapping and Annotation module employs the multi modality representation learning techniques to continuously refine its understanding of the contextual information within the interface, $W$ the node vectors and the relationships within it. This iterative learning process ensures that the system's mappings and annotations remain accurate and relevant, even as the underlying data changes.

$$S_n = \{M'_i \mid i \in W\} \quad (25)$$
$$S_n = T_{map}(S_e) \quad (26)$$

where,
$$T_{map}: M_i \mapsto M'_i \quad (27)$$
$$M'_i = f_{YOLO, OCR, VM}(M_i) \quad (28)$$

This module is fundamental to the cognitive capabilities of the DoRA framework, as it enables the system to construct a coherent and adaptable representation of the data landscape. By continually learning and updating its mappings and annotations, the system can maintain a high level of accuracy and efficiency in its automation tasks.

$$G = \{S_n\} \quad (29)$$

*C. Graph Aided Heuristic Search*

The Graph-aided Heuristic Search component of the DORA framework is designed to leverage the structured representation of data provided by the Learnable Mapping and Annotation module to efficiently navigate and retrieve relevant information. This search mechanism utilizes heuristic algorithms, which are guided by the graph structure and the normalized scores assigned to different data elements, to prioritize the most promising paths and minimize the search space.

The use of normalized scores is a key feature of this module, as it allows for a standardized comparison of different data points based on their relevance or importance to the task at hand. These scores can be derived from various factors, such as the frequency of occurrence, the strength of relationships, or the relevance to the user's query. By assigning scores to nodes and edges in the graph, the system can quickly identify the most pertinent information and focus its search efforts accordingly.

The heuristic aspect of the search algorithm is crucial for its efficiency, as it enables the system to make informed decisions about which paths to explore based on the available information and the current context. This approach reduces the computational overhead associated with exhaustive search methods and ensures that the system can retrieve relevant information in a timely manner.

The Graph-aided Heuristic Search module is an essential component of the DoRA framework, as it directly impacts the system's ability to quickly and accurately access the information required for various automation tasks. By optimizing the search process using heuristic and normalized scores, this module enhances the overall performance and effectiveness of the cognitive automation system.

*D. Knowledge Graph Augmented Language Modelling*

The Knowledge Graph Augmented Language Modelling component of the DoRA framework represents a significant advancement in natural language processing and dialogue generation. This module integrates the structured information from knowledge graphs into the language modelling process, enabling the system to generate more contextually relevant and coherent responses in conversational interfaces.

Grounding and context retrieval from subgraphs are key aspects of this module. By leveraging the connections and relationships encoded in the knowledge graph, the system can ensure that its responses are grounded in the relevant context, providing more accurate and informative node selections.

The integration of knowledge graphs into language modelling also facilitates the generation of knowledge-grounded dialogue, where the system references sub-graphs and nodes filtered in the graph aided search methods for substantiated response generation. By grounding language generation in the rich context provided by knowledge graphs, this module ensures that the optimal node selection is relevant and informative for consumption by AUTONODE.

---

**Algorithm 4** Knowledge Graph Augmented Language Modelling in the DoRA Framework

**Require:** Knowledge graph $G$, user query $Q$
**Ensure:** Contextually relevant and coherent response $R$
1: **function** RETRIEVESUBGRAPH($G$, $Q$)
2:    $S \leftarrow$ set of nodes in $G$ relevant to $Q$ using Graph-aided Search
3:    $SubG \leftarrow$ induced subgraph from $S$
4:    **return** $SubG$
5: **end function**
6: **function** GROUNDRESPONSE($SubG$, $Q$)
7:    $Context \leftarrow$ extract context from $SubG$
8:    $GroundedNodes \leftarrow$ select nodes from $SubG$ that best match $Q$
9:    **return** $Context, GroundedNodes$
10: **end function**
11: **function** GENERATERESPONSE($Context, GroundedNodes$)
12:    $Response \leftarrow$ initiate empty string
13:    **for** each node $n$ in $GroundedNodes$ **do**
14:      $Response \leftarrow Response +$ generate text grounded on $n$
15:    **end for**
16:    **return** $Response$
17: **end function**
18: $SubG \leftarrow$ RETRIEVESUBGRAPH($G, Q$)
19: $Context, GroundedNodes \leftarrow$ GROUNDRESPONSE($SubG, Q$)
20: $R \leftarrow$ GENERATERESPONSE($Context, GroundedNodes$)
21: **return** $R$

---

### E. From Language Modelling to Neuro-Symbolic Programming Paradigm (Appendix)

The transition from Language Modelling to a Neuro-Symbolic Programming Paradigm represents a paradigm shift in cognitive automation, as encapsulated in the DoRA framework. This shift involves integrating the flexibility and expressiveness of neural network-based language models with the structured reasoning capabilities of symbolic programming, creating a more holistic approach to cognitive automation.

Multimodal representation learning during training is a key aspect of this transition. By incorporating multiple data modalities, such as text, images, and site-graphs, into the learning process, the system can develop a more comprehensive understanding of the task at hand. This multimodal approach enables the system to enhance its cognitive capabilities, applying its learning to a wider range of automation tasks.

The Neuro-Symbolic Programming Paradigm offers several advantages over traditional language modelling approaches. By combining neural networks' ability to capture complex patterns and relationships with symbolic programming's logical reasoning and interpretability, the system can achieve a more nuanced and accurate understanding of the data. This integration enables the system to perform tasks that require both deep understanding and precise reasoning, such as natural language understanding, decision-making, and problem-solving.

The shift to a Neuro-Symbolic Programming Paradigm is a critical development in the DoRA framework, as it represents a significant step towards generalising cognitive automation across different domains and tasks. By leveraging multimodal representation learning and the synergies between neural networks and symbolic programming, the system can achieve a more advanced and versatile level of cognitive automation.

## V. SYSTEM ARCHITECTURE

In the preceding discourse, the integration of Process C alongside the DoRA module has been elucidated as the optimal architectural framework to address the cognitive challenges inherent in Robotic Process Automation (RPA). This architecture, characterized by the amalgamation of multiple components, evolves into a sophisticated system, entailing an intricate network of potential points of failure. Consequently, it becomes imperative to engineer the architecture of this system to not only embody scalability but also exhibit robustness, thereby ensuring its capacity to deliver efficient service to the user. This necessitates a meticulous design approach that prioritizes fault tolerance and adaptability, facilitating the system's ability to handle diverse and dynamic user demands effectively. The architecture of the proposed system incorporates an assemblage of sophisticated modules, including YOLO-V8, Optical Character Recognition (OCR), Retrieval-Augmented Generation (RAG), and an action-execution component. This composite structure underpins the operational efficacy of the system. The employment of the YOLO-V8 module is pivotal for the accurate detection of web elements within the interface, necessitating comprehensive training to fine-tune its performance for this specific application. Contrastingly, the integration of OCR models leverages their generic applicability, allowing for a seamless, one-time incorporation into the system. The action-execution module emerges as a critical entity within this framework, mandated to function with high reliability while ensuring scalability. It is imperative that this module operates in a non-blocking manner, facilitating asynchronous interactions with dynamically varying wait times. This capability is essential for addressing the challenges posed by web pages that exhibit delayed loading times. Incorporating the RAG module ushers in significant enhancements in terms of Turnaround Time (TAT) and system robustness. This is achieved through the module's capacity to catalog and recall the actions previously undertaken to fulfill specific objectives. Consequently, this enriches the system with the ability to efficiently execute tasks that are either identical or bear similarity to previously encountered objectives, thereby optimizing operational efficiency and adaptability.

## VI. EXPERIMENTATION AND RESULTS

In this section, we present the evaluation of our multimodal neuro-graphic retrieval agent framework through a combination of quantitative and qualitative experiments. Our primary goal is to assess the agent's performance and ability to operate on a diverse set of complex tasks across various applications.

### A. Experimental Setup

We evaluate the effectiveness of our agent (present and the former versions) against the SOTA multimodal agents using GPT-4 Vision and Gemini Pro, on tasks across different websites highlighting its adaptability and adoption in complex applications.

To comprehensively evaluate our methods we construct a benchmark which includes 5 web applications - **Apollo, Gmail, Calendar, Twitter and Contlo** encompassing more than 50 tasks (workflows), divided into three categories Level-1, Level-2 and Level-3 on the basis of their complexities and the number of steps involved for a workflow, each serving various purpose. In particular, to gain a more comprehensive insight into the proposed framework's approach involving the efficiency and the accuracy that the self-training module - DoRA imparts, we conduct an extensive case study on the Apollo website with over 50 standalone complex crowd sourced workflows. This case study allowed us to evaluate the framework's proficiency in handling complex tasks and automating RPA workflows which require cognitive involvement else wise.

The tasks and workflows are categorised on the basis of increasing complexity and the number of steps involved. **Level-1** (or L-1) tasks are relatively easier tasks/workflows with less than 5 steps involving static and intuitive User Interface within the websites / web applications. **Level-2** (or L-2) tasks, on the other hand are intermediately complex tasks with the number of steps ranging from 5-10. And **Level-3** (or L-3) tasks are the advanced and complex tasks with greater than 10 steps within the workflow. For the exploration and testing phase we capped the number of steps within the workflow to be 20.

B. *Results & Discussion*

Our comprehensive evaluation highlighted in Table 5.1 presents a comparative analysis of the successive versions of the AutoNode framework. With each iteration from v.1 to v.3, we observed a notable enhancement in the success rate, indicating a positive trend in the development and refinement of the AutoNode processes A, B, and C. This iterative process underscores the impact of incremental improvements and the importance of evolving architecture to achieve better performance.

TABLE I. COMPARISON OF AUTONODE VERSIONS

| Method | Architecture | Success Rate | |
|---|---|---|---|
| Process A | YOLO, OCR, GPT-4 Vision | 49.92 | - |
| Process B | * + Instruction Set, Verification | 70.58 | 76.46# |
| Process C | * + DoRA | 85.73 | 89.53# |

Table 5.1 Comparison of different versions of AutoNode. Success Rate refers to the average rate at which the agent framework completes task @first-pass. #Refers to the average success rate for more than 1 passes with verification loop incorporated. The scores are averaged over L-1, L-2 and L-3 tasks. *Refers to the addition to former architecture

TABLE II. EVALUATION RESULTS

| Agent | Accuracy | | |
|---|---|---|---|
| | Level 1 | Level 2 | Level 3 |
| Human* | 98.03 | 96.07 | 94.11 |
| MultiOn | 23.52 | 11.76 | - |
| SOC (HyperWrite) | 39.21 | 19.60 | - |
| AutoNode (w/o DoRA) | 84.31 | 68.62 | 59.82 |
| AutoNode | 92.15 | 90.19 | 86.27 |

Table 5.2 Comparison of different agents. Results reported on benchmark workflows with tasks categorized into Level 1(<5 steps), Level 2(5-10 steps) & Level 3(>10 steps) categories. The best multimodal agent performance is highlighted in contrast to the human performance* (accounting human error)

In Table 5.2, we extend our investigation to encompass a diverse array of agents, comparing their performance across three levels of task complexity. The human benchmark remains the gold standard, demonstrating superior performance with success rates consistently above 94% across all levels. Notably, the Human* agent exhibited only a modest decline in performance as task complexity increased from Level 1 to Level 3, which contrasts with the performance patterns observed in State of the Art multimodal agents and frameworks..

The Multi-On and Self Operating Computer (SOC)/Hyper-write agents delivered middling results, indicating that while they are capable of handling less complex tasks (Level 1), their performance significantly dwindles as they progress to Levels 2 and are unable to execute Level 3 tasks. This drop-off accentuates the challenge faced by current AI models in maintaining high performance with increasing task complexity.

Conversely, among the present open sourced agents and our proposed framework, the AUTONODE (with DoRA) showcased remarkable effectiveness, bridging the gap between human performance and automated systems, especially in higher complexity tasks (Level 3). The AUTONODE without the DoRA component presented the most significant variation in success rates between task levels. Its performance plummeted when transitioning from structured (Level 1) to more unstructured and complex tasks (Level 3), highlighting the pivotal role of the DoRA component in managing task complexity.

In conclusion, these results illustrate the critical importance of tailored process enhancements in automated agents to parallel human flexibility and efficiency. The experiments indicate potential pathways to augmenting agent's capabilities, specifically through modular improvements as seen with AutoNode, and stress the need for further research into exploration and adaptive mechanisms that allow AI to maintain high performance across varying levels of task complexity.

C. *Case Study ( Apollo Automation)*

To ascertain the efficacy and precision that the self-training module DoRA bestows, we carried out an in-depth case study within the Apollo automation environment. Our investigation delved into over 50 intricate, crowd-sourced workflows on the Apollo website, examining the AUTONODE framework's adeptness at executing complex tasks (L-2 and L-3) and automating RPA workflows that traditionally demand human cognitive skills. Additionally, the open-ended nature of the workflows' tasks allows us to assess the agent's problem-solving capabilities.

The AUTONODE model demonstrated a remarkable success rate, accurately completing 45 out of the 50 workflows in the first pass itself. For the remaining five workflows that presented unstructured challenges, the agent achieved a partial success rate varying between average accuracy ranging from **80% to 85%** in reaching a logical endpoint, despite not fully completing the entire workflow. The agent exhibited robust performance across all levels of complexity. This case study not only illuminated the agent's operational proficiency but also highlighted the robustness of DoRA in enhancing the framework's capability to tackle intricate workflows, thereby validating our approach in a real-world scenario. As can be seen from the comparison of results from the tables, our framework yields consistently better results than the SOTA multimodal agents and other open-sourced frameworks towards achieving a generalist agent for cognitive automation.

## VII. CONCLUSION

In the present manuscript, we have endeavored to investigate a feasible and prospective framework for cognitive Graphical User Interface (GUI) Automation. The empirical evidence derived from the conducted experiments attests to the robustness and reliability of the proposed framework. Future research directions include enhancing the Turnaround Time (TAT) for completing an objective. Currently, the framework requires approximately 10-15 minutes to execute an objective with a depth of 40. While the capacity to simultaneously run multiple sessions exists, it is deemed judicious to prioritize the refinement of the TAT and the overall reduction of the Waiting Time (WT) which includes future work on AutoRAGA and CogNAV architectures.

## VIII. ACKNOWLEDGEMENT